\documentclass[10pt]{article}
\usepackage{arxiv}
\usepackage{graphicx}

\usepackage{times}
\usepackage{latexsym}
\usepackage{microtype}

\usepackage{amsmath} 
\usepackage{amsfonts}

\usepackage{tabularx}
\usepackage{tabulary}
\usepackage{makecell}

\usepackage{multirow}
\usepackage{subcaption}
\usepackage{booktabs}
\usepackage{threeparttable}

\usepackage[inline, shortlabels]{enumitem}

\usepackage[acronym, shortcuts, nohypertypes={acronym}]{glossaries}
\newacronym{AER}{AER}{automatic emotion recognition}
\newacronym{ASR}{ASR}{automatic speech recognition}
\newacronym{BoW}{BoW}{bag-of-words}
\newacronym{BN}{BN}{batch normalisation}
\newacronym{BERT}{BERT}{Bidirectional Encoder Representations from Transformers}
\newacronym{bGRU}{bGRU}{bidirectional gated recurrent unit}
\newacronym{bLSTM}{bLSTM}{bidirectional long short-term memory}
\newacronym{CCC}{CCC}{concordance correlation coefficient}
\newacronym{CV}{CV}{cross-validation}
\newacronym{CNN}{CNN}{convolutional neural network}
\newacronym{DL}{DL}{deep learning}
\newacronym{DNN}{DNN}{deep neural network}
\newacronym{FCNN}{FCNN}{fully connected neural network}
\newacronym{FiLM}{FiLM}{feature-wise linear modulation}
\newacronym{HLD}{HLD}{higher-level descriptor}
\newacronym{LLD}{LLD}{low-level descriptor}
\newacronym{LOSO}{LOSO}{leave-one-speaker-out}
\newacronym{LSTM}{LSTM}{long short-term memory}
\newacronym{MAE}{MAE}{mean absolute error}
\newacronym{MFCC}{MFCC}{Mel frequency ceptral coefficient}
\newacronym{MSE}{MSE}{mean squared error}
\newacronym{ML}{ML}{machine learning}
\newacronym{MLP}{MLP}{multilayer perceptron}
\newacronym{NLP}{NLP}{natural language processing}
\newacronym{NLU}{NLU}{natural language understanding}
\newacronym{NOSS}{NOSS}{non-semantic speech benchmark}
\newacronym{NLLoss}{NLLoss}{non-negative likelihood loss}
\newacronym{PCC}{PCC}{Pearson correlation coefficient}
\newacronym{ReLU}{ReLU}{rectified linear unit}
\newacronym{SAM}{SAM}{self-assesment manikin}
\newacronym{SER}{SER}{speech emotion recognition}
\newacronym{SGD}{SGD}{stochastic gradient descent}
\newacronym{SVM}{SVM}{support vector machine}
\newacronym{TPR}{TPR}{the true positive rate}
\newacronym{UAR}{UAR}{unweighted average recall}
\newacronym{WA}{WA}{weighted accuracy}

\usepackage[natbib, bibencoding=utf8, citestyle=numeric, bibstyle=ieee, maxbibnames=999, maxcitenames=2, mincitenames=1, sortcites]{biblatex}
\bibliography{bibliography.bib}

\usepackage[capitalise, nameinlink, noabbrev]{cleveref}

\newcommand{\eg}{e.\,g. }

\title{
    Multistage linguistic conditioning of convolutional layers for speech emotion recognition
}

\author{Andreas~Triantafyllopoulos\textsuperscript{1,2},
        Uwe~Reichel\textsuperscript{1},
        Shuo~Liu\textsuperscript{2},
        Stephan~Huber\textsuperscript{1},
        Florian~Eyben\textsuperscript{1},
        and~Bj\"orn~W.~Schuller\textsuperscript{1,2,3}\\
\textsuperscript{1} audEERING GmbH, Gilching, Germany\\
\textsuperscript{2} EIHW, University of Augsburg, Augsburg, Germany\\
\textsuperscript{3} GLAM, Imperial College, London, UK\\
}

\begin{document}

\maketitle

\begin{abstract}
In this contribution, we investigate the effectiveness of deep fusion of text and audio features for categorical and dimensional speech emotion recognition (SER).
We propose a novel, multistage fusion method where the two information streams are integrated in several layers of a deep neural network (DNN), and contrast it with a single-stage one where the streams are merged in a single point.
Both methods depend on extracting summary linguistic embeddings from a pre-trained BERT model, and conditioning one or more intermediate representations of a convolutional model operating on log-Mel spectrograms.
Experiments on the MSP-Podcast and IEMOCAP datasets demonstrate that the two fusion methods clearly outperform a shallow (late) fusion baseline and their unimodal constituents, both in terms of quantitative performance and qualitative behaviour.
Overall, our multistage fusion shows better quantitative performance, surpassing alternatives on most of our evaluations.
This illustrates the potential of multistage fusion in better assimilating text and audio information.
\end{abstract}





\glsresetall
\section{Introduction}
\label{sec:introduction}
\Ac{AER} is an important component of human-computer interfaces, with applications in health and wellbeing, multimedia information retrieval, and dialogue systems.
Human emotions are expressed in, and can accordingly be identified from, different modalities, such as speech, gestures, and facial expressions~\citep{zeng2008survey, calvo2010affect}.
Over the years, different modalities have proven more conducive to the recognition of different emotional states.
For example, video has shown better performance at valence recognition, whereas acoustic cues are better indicators of arousal~\citep{calvo2010affect}.
A plethora of previous works have thus investigated different ways to improve \ac{AER} by combining several information streams: \eg audio, video, text, gestures, and physiological signals.
Underlying these computational approaches are distinct emotion theories, with the two most commonly-used ones being \emph{discrete (or basic) emotion theories}~\citep{ekman1992argument} and \emph{dimensional} ones~\citep{russell1977evidence}.

In the present contribution, we are primarily interested in \ac{SER}.
Speech, as the primary mode of human communication, is also a major conduit of emotional expression~\citep{calvo2010affect}.
Accordingly, \ac{SER} has been a prominent research area in affective computing for several years~\citep{schuller2018speech}.
Although it constitutes a single modality, it is often analysed as two separate information streams, a \emph{linguistic} (\emph{what} has been said) and a \emph{paralinguistic} one (\emph{how} it has been said)~\citep{calvo2010affect, schuller2018speech, atmaja2022survey}.

However, the two streams are not independent.
Previous studies have established that the interaction between acoustic descriptors and emotional states depends on the linguistic content of an utterance~\citep{scherer1984vocal}.
Moreover, text information is better suited for valence and audio for arousal recognition~\citep{calvo2010affect}.
As a result, several works have attempted to more tightly integrate the two streams in an attempt to model their complex interrelationship and obtain more reliable recognition performance~\citep{zeng2008survey, calvo2010affect, atmaja2022survey}.

Recently, deep fusion architectures have proven very successful at utilising linguistic and acoustic cues for \ac{SER}~\citep{siriwardhana2020jointly, lee2018convolutional, chen2019complementary, georgiou2019deep}.
However, most existing such methods are primarily based on sequential models (\eg \acp{LSTM}) operating on expert-based acoustic descriptors~\citep{lee2018convolutional, chen2019complementary}.
Whereas such methods have a long history in the field of \ac{SER}, in recent years \acp{CNN} operating on raw audio or low-level features have been shown capable of learning good representations that lead to better performance~\citep{trigeorgis2016adieu, fayek2017evaluating, neumann2017attentive}.
Thus, the combination of multistage fusion with the representation power of \acp{CNN} for auditory processing is a natural next step in the attempt to closely integrate the acoustic and linguistic information streams. 

Inspired by such works, we propose a novel, \ac{CNN}-based multistage fusion architecture where summary linguistic embeddings extracted from a pre-trained language model are used to condition multiple intermediate layers of a \ac{CNN} operating on log-Mel spectrograms.
We contrast it with a single-stage \ac{DNN} architecture, where the two information streams are fused at a single point, and a late fusion method, where unimodal models are trained independently and their decisions are aggregated.

The rest of this contribution is organised as follows.
We begin by presenting an overview of related works in \cref{sec:related}, with an emphasis on single-stage and multistage fusion approaches.
We then describe our architecture in \cref{sec:method} and experimental settings in \cref{sec:setup}.
Results are reported in \cref{sec:categories} and \cref{sec:dimensions} for categories and dimensions, respectively.
We end with a conclusion in \cref{sec:conclusion}.

\section{Related work}
\label{sec:related}

In this section, we present an overview of related works.
We begin by a brief overview of the state-of-the-art in unimodal, either text- or acoustic-based, \ac{AER} methods.
We then present an overview of fusion methods.
For a more in-depth, recent survey on bimodal \ac{SER}, we refer the reader to \citet{atmaja2022survey}.

\subsection{Unimodal Speech Emotion Recognition}
The goal of audio-based \ac{SER} systems is to estimate the target speaker's emotion by analysing their voice~\citep{schuller2018speech}.
For paralinguistics, this is traditionally handled by extracting a set of \acp{LLD}, such as \acp{MFCC} or pitch, which capture relevant cues~\citep{schuller2013interspeech}.
In recent years, there has been a rise in deep learning architectures based on CNNs~\citep{trigeorgis2016adieu} or transformers~\citep{wagner2022dawn} which outperform the more traditional approaches.

Early works on text-based emotion recognition utilised affective lexica, such as the WordNet-Affect dictionary~\citep{strapparava2004wordnet}, to generate word-level scores, which could then be combined using expert rules to derive a sentence-level prediction~\citep{perikos2013recognizing}.
In recent years, the use of learnt textual representations, like word2vec~\citep{mikolov2013distributed}, has substituted these methods.
Here too, attention-based models (transformers)~\citep{vaswani2017attention} have shown exceptional performance on several \ac{NLP} tasks.
These models are usually pre-trained on large, unlabelled corpora using some proxy task, \eg masked language prediction~\citep{devlin2018bert}, which enables them to learn generic text representations.

\subsection{Bimodal Speech Emotion Recognition}

Early work in multimodal fusion has primarily followed the shallow fusion paradigm\,\citep{atrey2010multimodal, poria2017review}.
Several early systems depended on hand-crafted features, usually \acp{HLD}, extracted independently for each modality, which were subsequently processed by a fusion architecture adhering to the early or late fusion paradigm.
Early fusion corresponds to feeding the \acp{HLD} from both information streams as input to a single classifier.
Late fusion, on the other hand, is achieved by training unimodal classifiers independently, and subsequently aggregating their predictions.
The aggregation can consist of simple rules (\eg  averaging the predictions) or be delegated to a cascade classifier~\citep{steidl2009hinterland}.

With the advent of \ac{DL}, traditional, shallow fusion methods have been substituted by end-to-end multimodal systems\,\citep{tzirakis2017end} where the different modalities are processed by jointly-trained modules.
We differentiate between \emph{single-stage} and \emph{multistage} fusion.
Single-stage fusion is the natural extension of shallow fusion methods, where the different modalities are first processed separately by independent differentiable modules~\citep{chen2019complementary, priyasad2020attention}.
Although these methods have consistently outperformed both the unimodal baselines and shallow fusion alternatives, they nevertheless build on independently learnt unimodal representations constructed by modules agnostic to the presence of other modalities.
In an attempt to utilise the power of \ac{DL} to learn useful representations after several layers of processing, the community has also pursued multistage fusion paradigms, where the processing of different modalities is intertwined in multiple layers of a \ac{DNN}~\citep{georgiou2019deep,zadeh2018multimodal,tseng2021multimodal}.

Our proposed multistage fusion mechanism draws inspiration from recent approaches in denoising~\citep{keren2018scaling, gfeller2020one} and speaker adaptation~\citep{triantafyllopoulos2021deep}.
In general, all these approaches utilise two \acp{DNN}: one devoted to the primary task, and a second providing additional information through some fusion mechanism.
The two networks can be trained independently or jointly.
The fusion mechanisms utilised in these works all operate on the same principle: the embeddings produced by the secondary network modulate the output of several (usually all) convolution layers of the primary network either by shifting or a combination of scaling and shifting.

\section{Proposed Method}
\label{sec:method}
\begin{figure*}[!t]
    \centering
    \hspace*{-0.5cm}
    \begin{subfigure}{0.47\textwidth}
        \includegraphics[width=\textwidth]{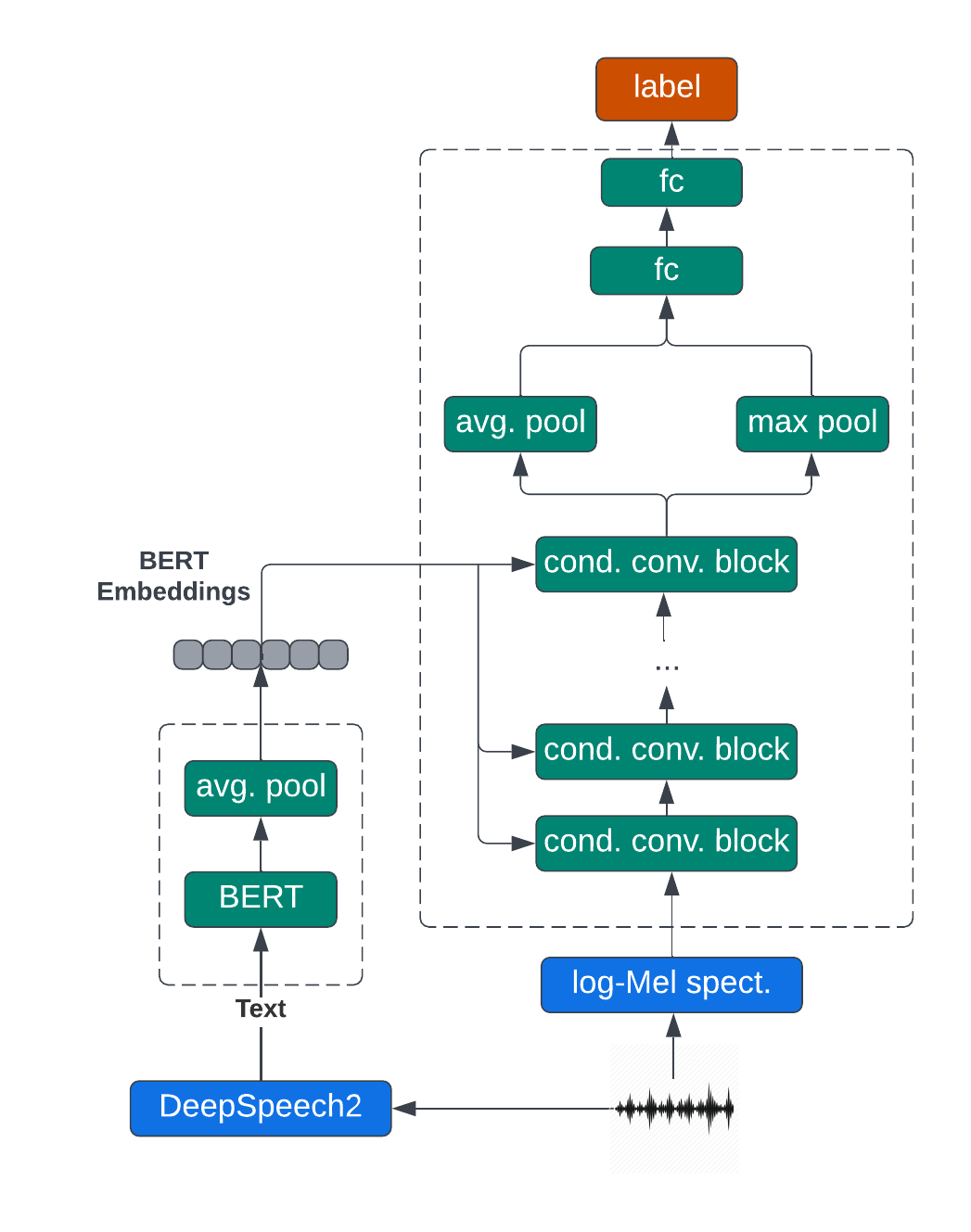}
        \vspace*{-1.2cm}
        \caption{\textbf{MFCNN14}}
        \label{subfig:MFCNN14}
    \end{subfigure}~%
    \hspace*{0.2cm}
    \begin{subfigure}{0.47\textwidth}
        \includegraphics[width=\textwidth]{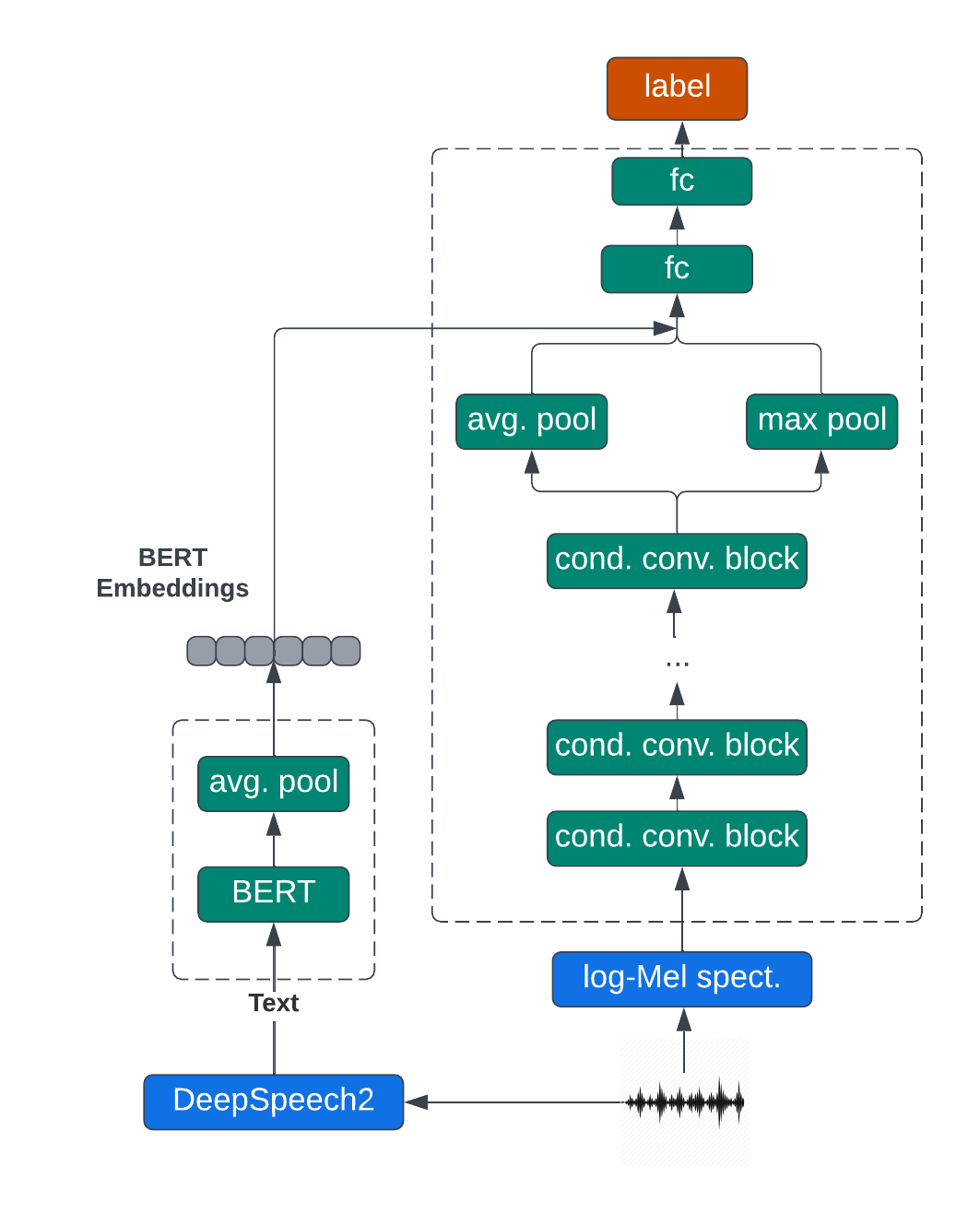}
        \vspace*{-1.2cm}
        \caption{\textbf{SFCNN14}}
        \label{subfig:SFCNN14}
    \end{subfigure}
    \caption{
    Diagrams of the architectures used in this work illustrating the processing of a single utterance.
    (a) \textbf{MFCNN14: Multistage fusion architecture}. 
    (b) \textbf{SFCNN14: Single-stage fusion architecture}. 
    The same architecture was used for the baseline CNN14, but without the fusion.
    }
    \vspace{-0.2cm}
    \label{fig:arch}
\end{figure*}

The focus of the current contribution is on tightly integrating acoustic and linguistic information.
This is achieved by proposing a multistage fusion approach where linguistic embeddings condition several intermediate layers of a \ac{CNN} processing audio information.
The overall architecture comprises of two constituent networks: a (pre-trained) text-based model that provides linguistic embeddings and an auditory \ac{CNN} whose intermediate representations are conditioned on those embeddings.

The two unimodal architectures can obviously be trained independently for emotion recognition and their predictions can be aggregated in a straightforward way (\eg by averaging) to produce a combined output.
This simple setup is used in some of our experiments as a shallow (late) fusion baseline.
Additionally, the two information streams can be combined in single-stage fashion, with the linguistic embeddings fused at a single point with the embeddings generated by the \ac{CNN}; a setup which forms a deep fusion baseline with which to compare our method.

As our unimodal text model, we use BERT~\citep{devlin2018bert}, a pre-trained model with a strong track-record on several \ac{NLP} tasks.
As our baseline acoustic model, we use the CNN14 architecture introduced by \citet{kong2019panns}, which was found to give good performance for emotional dimensions~\citep{triantafyllopoulos2021role}.
CNN14 consists of $12$ convolution layers and $2$ linear ones, with mean and max pooling following the last convolution layer to aggregate its information over the time and frequency axes.
These unimodal networks form the building blocks of our fusion methods, which are illustrated in \cref{fig:arch}.
Both architectures first pass the text input through a pre-trained BERT model, and then use it to condition one or more layers of the CNN14 base architecture.

Our proposed multistage fusion method, shown in \cref{subfig:MFCNN14}, relies on fusing linguistic representations in the form of embeddings extracted from a pre-trained BERT model with the intermediate layer outputs of an acoustics-based \ac{CNN} model.
Linguistic embeddings are computed by averaging the token-embeddings returned by BERT for each utterance.
Similar to prior works~\citep{keren2018scaling, triantafyllopoulos2021deep}, we use the averaged embeddings to shift the intermediate representations of each block.
Given an input $\pmb{X} \in \mathbb{R}^{T_{\text{in}} \times F_{\text{in}}}$ to each convolution block,
with $T$ and $F$ being the number of time windows and frequency bins, respectively, and the average BERT embeddings $\pmb{E}_\text{L} \in \mathbb{R}^{\text{L}_\text{dim}}$, the output, $\pmb{Y} \in \mathbb{R}^{T_{\text{out}} \times F_{\text{out}}}$, is computed as follows:
\begin{align}
    \pmb{H}_1 =& \text{ReLU}(\text{BN}(\text{CONV}(\pmb{X})))\text{,}\\
    \pmb{H}_2 =& \text{ReLU}(\text{BN}(\text{CONV}(\pmb{H}_1)))\text{,}\\
    \pmb{H_3} =& \text{MaxPool}(\pmb{H}_2)\text{, and}\\
    \pmb{Y} =& \pmb{H}_3 + \text{PROJ}(\pmb{E}_\text{L})\text{,}
\end{align}
where ReLU stands for the rectified linear unit activation function~\citep{nair2010rectified}, BN for batch normalisation~\citep{ioffe2015batch}, and CONV for 2D convolutions.
The projection ($\text{PROJ}$) is implemented as a trainable linear layer which projects the input embeddings $\pmb{E}_\text{L}$ to a vector, $\pmb{E}_\text{P} \in \mathbb{R}^{F_{\text{out}}}$, with the same dimensionality as the output feature maps:
\begin{equation}
    \pmb{E}_\text{P} = \pmb{W} \times \pmb{E}_\text{L} + \pmb{b} \text{,}
\end{equation}
where $\pmb{W}$ and $\pmb{b}$ are the trainable weight and bias terms, respectively.
Thus, this conditioning mechanism is tantamount to adding a unique bias term to each output feature map of each convolution block.

The single-stage fusion architecture integrates acoustics and linguistics at a single point: immediately after the output of the last CNN14 convolution layer.
The linguistic embeddings are first projected to the appropriate dimension, and then added to the acoustic representations produced by the convolution network.
The shallow fusion architecture is shown in \cref{subfig:SFCNN14}.

\section{Experimental setup}
\label{sec:setup}
\subsection{Datasets}
\label{subsec:data}
\subsubsection{IEMOCAP}
We use the Interactive Emotional Dyadic Motion Capture (IEMOCAP) dataset \citep{busso2008iemocap}, a multimodal emotion recognition dataset collected from $5$ pairs of actors, each acting a set of scripted and improvised conversations, resulting in a total of $10,039$ utterances.
It has been annotated for the emotional dimensions of \emph{arousal}, \emph{valence}, and \emph{dominance} on a $5$-point Likert scale, with individual ratings averaged over all annotators to produce the gold standard.
It has also been annotated for the emotion categories of \emph{neutral}, \emph{excited}, \emph{surprised}, \emph{happy}, \emph{frustrated}, \emph{sad}, \emph{angry}, \emph{afraid}, and \emph{disgusted}.
The dataset additionally contains gold standard transcriptions, which we use in our experiments.
As IEMOCAP does not contain official train/dev/test splits, we follow the established convention of evaluating using \ac{LOSO} \ac{CV}\,\citep{tseng2021multimodal,priyasad2020attention,poria2018multimodal,siddique2020multitask}, where we use all utterances of each speaker once as the test set, each time using the utterances of their pair as the validation set, resulting in a total of $10$ folds.

\subsubsection{MSP-Podcast}
MSP-Podcast\,\citep{lotfian2019msppodcast} is a recently-introduced data set for \ac{SER}.
The dataset is constantly growing and new releases are made every year; we used version v1.7, which was the latest one available to us for our experiments.
It is split into speaker independent partitions, with a training set consisting of $38\,179$ segments, a development set of $7\,538$ segments, collected from 44 speakers (22 male -- 22 female), and a $12\,902$ segment test set, consisting of 60 speakers (30 male -- 30 female).
The dataset has been annotated for the emotional dimensions of \textit{arousal}, \textit{valence}, \textit{dominance}, as well as for the emotion categories of \textit{angry}, \textit{contemptful}, \textit{disgusted}, \textit{afraid}, \textit{happy}, \textit{neutral}, \textit{sad}, and \textit{surprised}.
The emotional dimensions have been annotated on a 7-point Likert scale on the utterance level, and scores by individual annotators have been averaged to obtain a consensus vote.
All experiments on MSP-Podcast are performed on the official train/dev/test splits.

As we have no ground truth transcriptions for MSP-Podcast, we generated them automatically using an open-source implementation of DeepSpeech2\,\citep{amodei2016deep}\footnote{https://github.com/mozilla/DeepSpeech}.
Whereas other works have used more advanced, proprietary models~\citep{pepino2020fusion}, we opted for a widely-used open-source alternative for reproducibility.
However, this model achieves a worse \ac{ASR} performance than that of proprietary models.
As it has been shown by several previous works that the performance of text-based and fusion approaches improves with better \ac{ASR} models~\citep{yoon2018multimodal, sahu2019multi}, we also expect our method to yield correspondingly better results, and do not consider this a critical limitation of our work.

\subsection{Experimental procedure}
As discussed in \cref{sec:introduction}, we consider both a categorical and a dimensional model of emotion.
For discrete emotion recognition, most works on the two datasets considered here pursue a 4-class classification problem, utilising the emotion classes of \emph{\{angry, happy, neutral, sad\}}, while further fusing the emotion class of \emph{excited} with that of \emph{happy} for IEMOCAP~\citep{atmaja2022survey}.
To make our results comparable, we follow this formulation as well.
For these experiments, we report \ac{UAR}(\%), the standard evaluation metric for this task which also accounts for class imbalance, and additionally show confusion matrices.
To mitigate the effect of class imbalance, which
is particularly pronounced for MSP-Podcast, we use a weighted variant of cross-entropy, where the loss for each sample is weighted by the inverse frequency of the class it belongs to.
For dimensional \ac{SER}, where we have continuous values for the dimensions of arousal, valence and dominance, we formulate our problem as a standard regression task and evaluate based on \ac{CCC} --the standard evaluation metric for dimensional emotion~\citep{parthasarathy2017jointly, li2021contrastive}-- and also train with the \ac{CCC} loss~\citep{parthasarathy2017jointly, li2021contrastive}.


However, multi-tasking potentially entangles the three dimensions and therefore complicates our analysis.
Moreover, whereas the \ac{CCC} loss is widely used for emotional dimension modelling, it is not the standard loss for regression tasks.
Thus, we begin by considering single-task models trained with the standard \ac{MSE} loss.

We perform our experiments by separately training on IEMOCAP and MSP-Podcast.
As mentioned, we perform $10$-fold \ac{LOSO} \ac{CV} for the first and use the official train/dev/test partitions for the latter.
We also report cross-domain results.
Cross-domain results are obtained by evaluating models trained with one dataset on the other.
For MSP-Podcast, where a single model is trained, we evaluate it on the entire IEMOCAP dataset.
For IEMOCAP, where $10$ models are trained for each experiment, we evaluate all $10$ of them on the test set of MSP-Podcast, and compute the average performance metric for each task.
As the emotional dimensions of the two datasets are annotated with different scales ($5$-point scale for IEMOCAP and $7$-point scale for MSP-Podcast), we evaluate cross-corpus performance using \ac{PCC} instead of \ac{CCC}, as the former is unaffected by differences in the scale.

For each dataset and task, we thus always perform the following experiments:
\begin{itemize}[topsep=0.5pt,itemsep=1pt,partopsep=1pt, parsep=1pt]
    \item \textbf{CNN14}: the unimodal, acoustics-only baseline,
    \item \textbf{SFCNN14}: our single-stage fusion architecture,
    \item \textbf{MFCNN14}: our multistage fusion architecture.
\end{itemize}
All models are trained for $60$ epochs with a learning rate of $0.01$ and a batch size of $64$ using \ac{SGD} with a Nesterov momentum of $0.9$.
We select the model that performed best on each respective validation set.
In order to avoid statistical fluctuations due to random seeds, we run each experiment $5$ times and report mean and standard deviation.

Additionally, for some configurations, we fine-tune the pre-trained BERT model, a practice which has recently emerged as the standard linguistic baseline showing strong performance on several \ac{NLP} tasks.
This particular configuration is not considered in all our experiments, as the number of different formulations examined made the computational cost of fine-tuning several BERT models prohibitive.
Specifically, we train $5$ models with different random seeds for categorical emotion recognition and multi-tasking on the emotional dimensions of MSP-Podcast.
We omit single-task experiments.

As pre-trained model, we selected {\em bert-base-uncased} distributed by Huggingface%
\footnote{https://huggingface.co/bert-base-uncased} with a final linear layer.
As in the other experiments, we use the weighted cross-entropy loss for classification, the \ac{MSE} loss for single task regression, and the mean \ac{CCC} loss averaged over all targets for multitask regression. 
For all conditions, the maximum token length was set to $128$, and the batch size to $32$. 
For fine-tuning, we chose the Adam optimiser with fixed weight decay (learning rate: $2e-5$, betas: $0.9$ and $0.999$, epsilon: $1e-06$, weight decay: $0.0$, no bias correction), and a linear schedule with $1000$ total and $100$ warmup steps. 
We trained for $4$ epochs with early stopping based on a \ac{UAR} or \ac{CCC} decrease on the development set for classification and regression, respectively.

\section{Emotional categories}
\label{sec:categories}
\begin{table*}[t]
    \caption{
    \Ac{UAR}(\%) in- and cross-domain results for 4-class emotion recognition (chance level: $25\,\%$).
    MSP-Podcast in-domain results computed on the official test set, whereas IEMOCAP in-domain results correspond to \acl{LOSO} \acl{CV}, where data from each speaker is used once as the test set and the other speaker in their session is used as the development set.
    Cross-domain results are reported on the official test set for MSP-Podcast and the entire dataset for IEMOCAP.
    Average (and standard deviation) results computed over $5$ runs.
    Fusion results (SFCNN14 and MFCNN14) that are significantly different than the unimodal baselines (CNN14 and BERT) as determined by two-sided independent t-tests ($p < 0.05$) are marked by $^*$ and $^\dag$, respectively.
	}
	\label{table:emotion}
	\centering
    \begin{tabular}{c|ll|ll}
    \toprule
    \textbf{Train} & \multicolumn{2}{c|}{\textbf{MSP-Podcast}} & \multicolumn{2}{c}{\textbf{IEMOCAP}}\\
    \midrule
    \textbf{Test} & \textbf{MSP-Podcast} & \textbf{IEMOCAP} & \textbf{IEMOCAP} & \textbf{MSP-Podcast}\\
    \midrule
    & \textbf{\ac{UAR}} & \textbf{\ac{UAR}} & \textbf{\ac{UAR}} & \textbf{\ac{UAR}}\\
    \midrule
    \textbf{BERT}    & 48.9 (0.4) & 45.0 (0.7) & 71.1 (0.2) & \textbf{40.5 (0.1)} \\
    \textbf{CNN14}   & 48.3 (0.6) & 41.3 (2.9) & 55.8 (1.0) & 31.5 (1.9)\\
    \textbf{SFCNN14} & \textbf{56.0 (1.3)}$^{*\dag}$ & \textbf{46.4 (3.9)} & 64.1 (0.7)$^{*\dag}$ & 33.3 (1.8)$^{*\dag}$\\
    \textbf{MFCNN14} & 54.2 (0.8)$^{*\dag}$ & \textbf{46.4 (2.0)}$^*$ & \textbf{72.6 (0.7)$^{*\dag}$} & 35.6 (2.1)$^{*\dag}$\\
    \bottomrule
    \end{tabular}
\end{table*}
\begin{figure}[!t]
    \centering
    \begin{subfigure}{0.49\textwidth}
    \includegraphics[width=\linewidth]{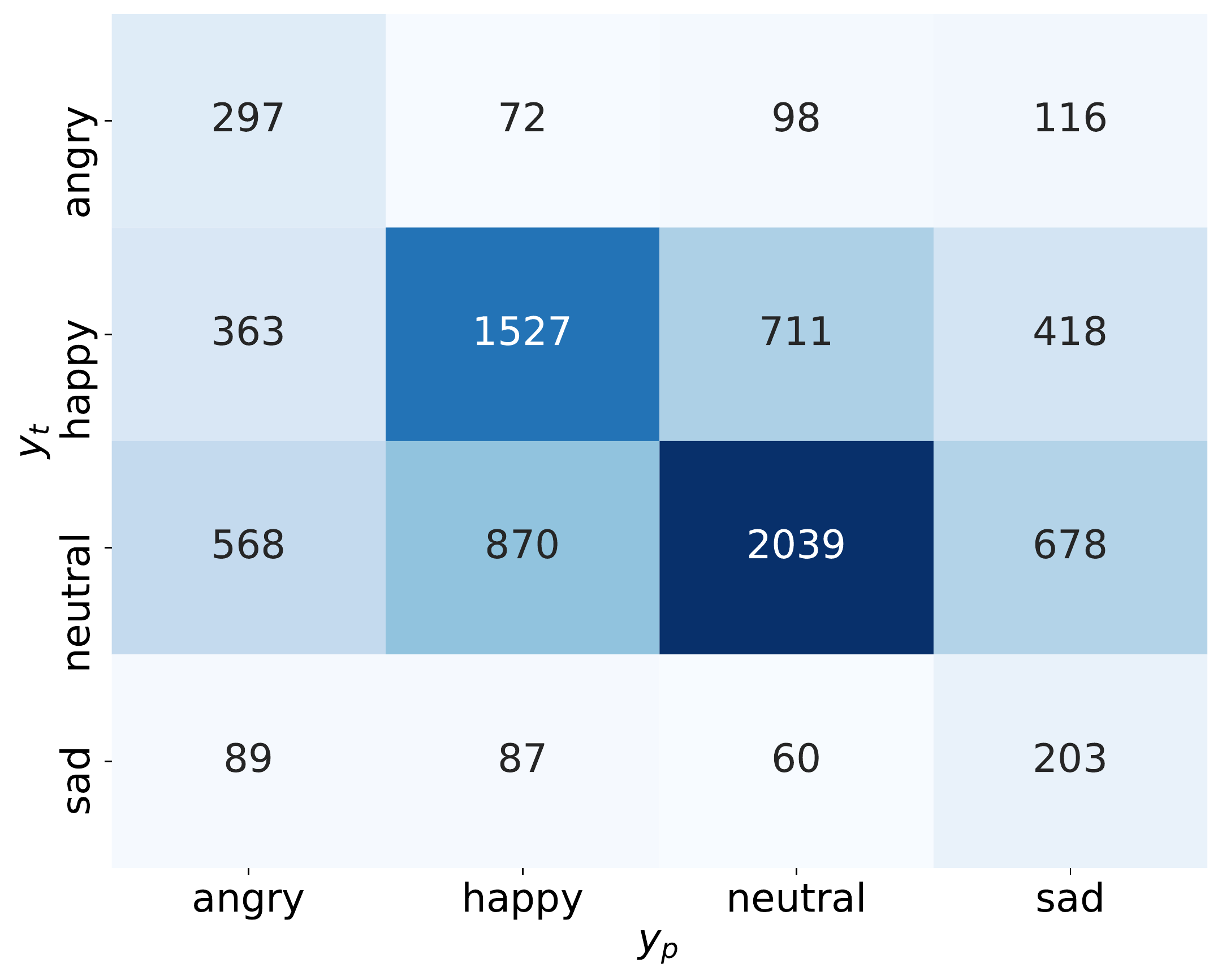}
    \caption{BERT}
    \end{subfigure}~%
    \begin{subfigure}{0.49\textwidth}
    \includegraphics[width=\linewidth]{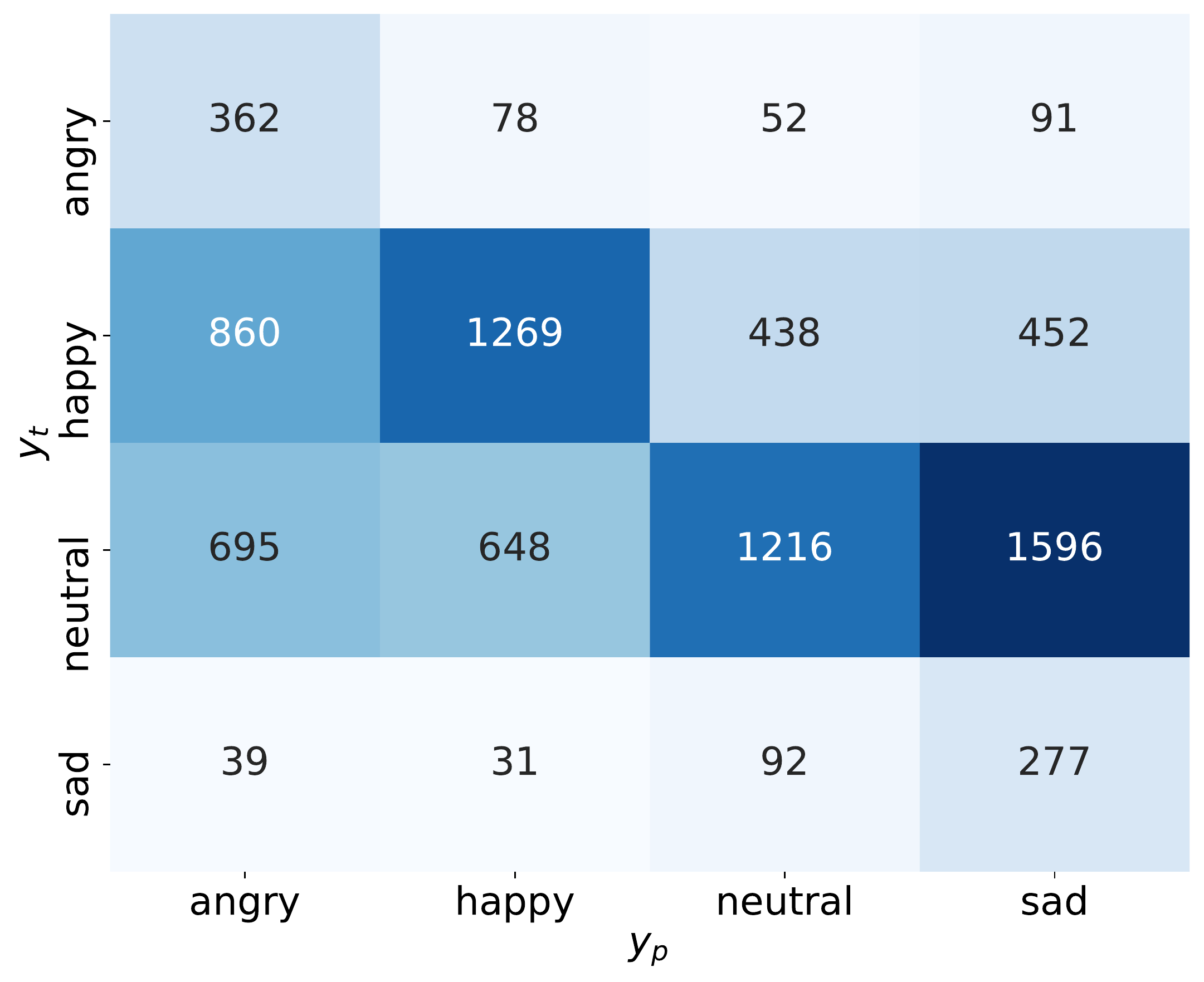}
    \caption{CNN14}
    \end{subfigure}
    \begin{subfigure}{0.49\textwidth}
    \includegraphics[width=\linewidth]{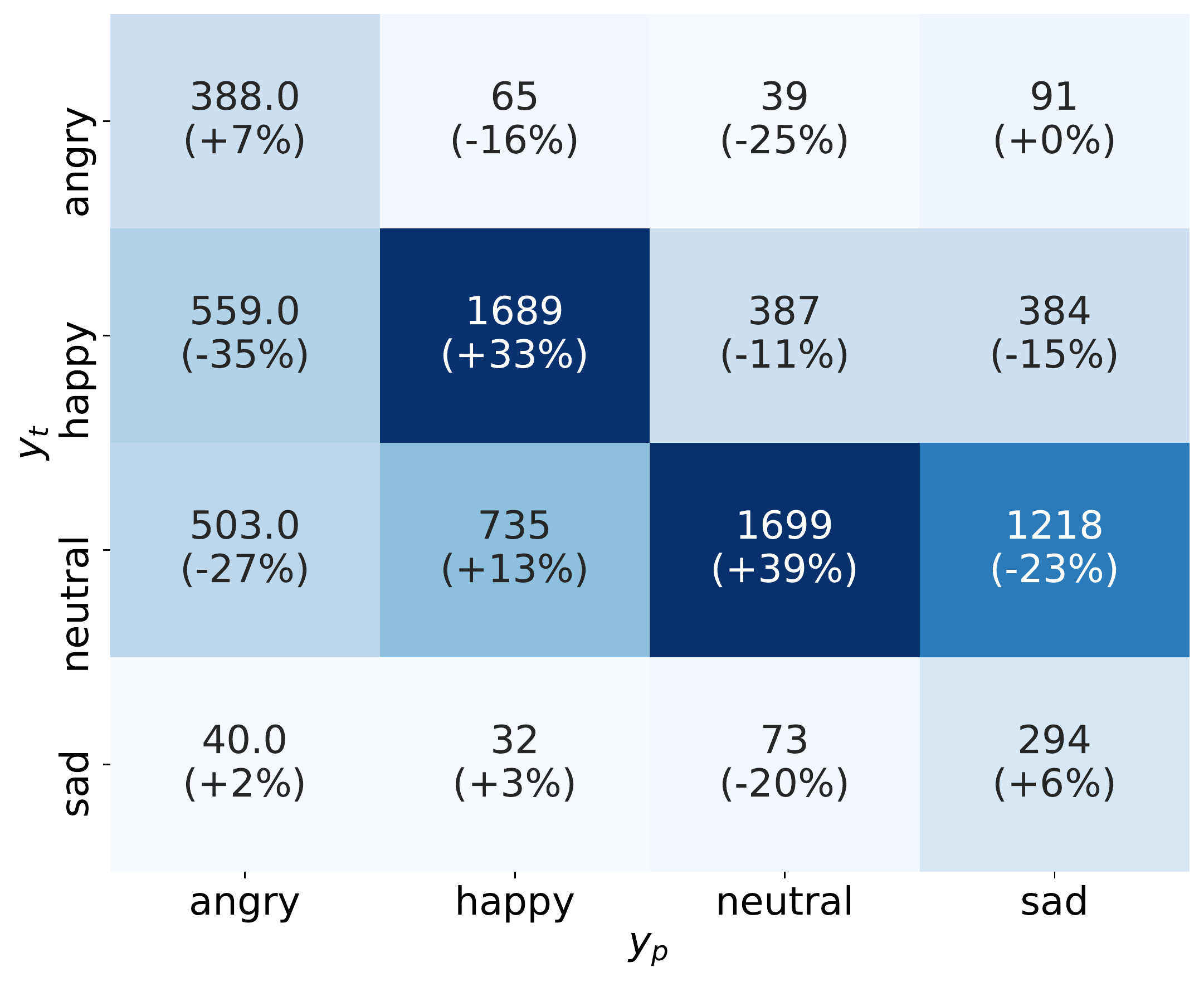}
    \caption{SFCNN14}
    \end{subfigure}~%
    \begin{subfigure}{0.49\textwidth}
    \includegraphics[width=\linewidth]{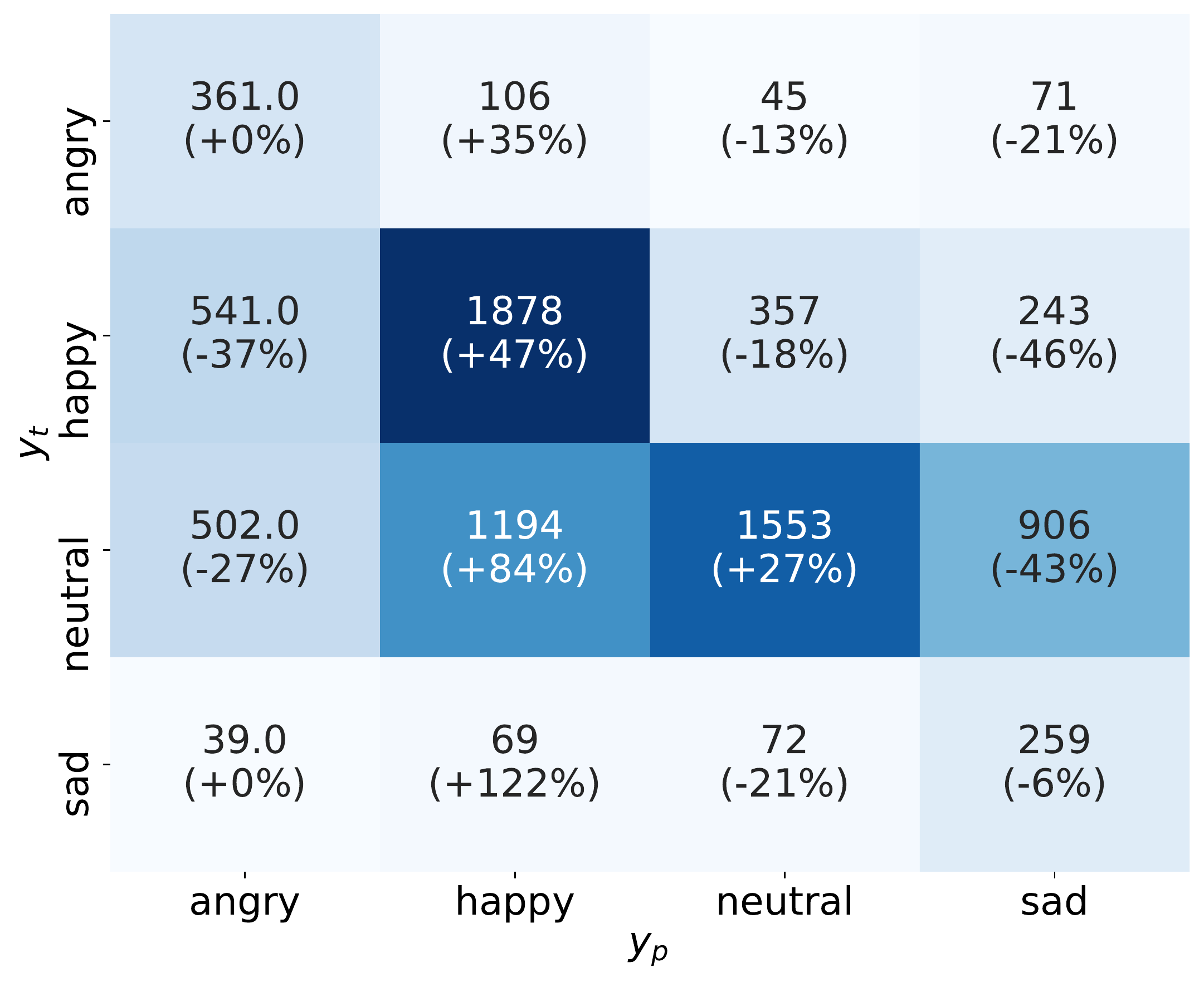}
    \caption{MFCNN14}
    \end{subfigure}
    \caption{
    Confusion matrices for the 4-class emotion classification task on MSP-Podcast.
    For each approach, we show results for the best performing seed.
    For the two fusion models, we additionally show $\%$ change with respect to CNN14 for easier comparison.
    }
    \label{fig:podcast_confusions}
\end{figure}

We begin by considering the 4-class emotion classification problem discussed in \cref{sec:setup}.
\cref{table:emotion} presents in- and cross-domain results on MSP-Podcast and IEMOCAP for both unimodal baselines and both fusion methods.
Interestingly, CNN14 performs worse than BERT on both MSP-Podcast, with a \ac{UAR} of $48.3\,\%$ vs $48.9\,\%$, and on IEMOCAP ($55.8\,\%$ vs $71.1\,\%$), while also achieving the best cross-corpus performance when training on IEMOCAP and evaluating on MSP-Podcast.
This shows that linguistics carry more emotional information on both datasets, but more so for IEMOCAP.
On the one hand, this could be due to the noisy transcriptions used for MSP-Podcast.
On the other hand, this dataset is more naturalistic than IEMOCAP, where actors could have relied more on text for conveying their emotions, especially in the case of scripted conversations.

Bimodal fusion leads to consistently higher performance compared to CNN14.
Both architectures perform significantly better than the unimodal audio baseline for both datasets, with SFCNN14 performing slightly better on MSP-Podcast, and MFCNN14 considerably outperforming it on IEMOCAP.
Moreover, in the case of IEMOCAP, only MFCNN14 is better than BERT, whereas SFCNN14 is significantly worse than it.
In terms of cross-corpus results, the two fusion models yield the same performance when trained on MSP-Podcast and tested on IEMOCAP, with MFCNN14 being better on the reverse setup.
In both cases, performance is severely degraded, which illustrates once more the challenges associated with cross-corpus \ac{AER}.
Interestingly, the degradation is lower for models trained on MSP-Podcast, indicating that larger, naturalistic corpora can lead to overall better generalisation.
Thus, while SFCNN14 outperforms MFCNN14 on MSP-Podcast, the latter is showing better generalisation capabilities and better performance on IEMOCAP, indicating that this form of fusion has great potential.



\cref{fig:podcast_confusions} additionally shows the confusion matrices for the best performing models (based on the validation set) on the MSP-Podcast test set.
For both CNN14 and BERT, we observe poor performance, with large off-diagonal entries.
Notably, BERT is better at recognising \emph{happy} and \emph{neutral} while CNN14 is better for \emph{angry} and \emph{sad}.
For CNN14, the most frequent misclassifications occur when \emph{happy} is misclassified into \emph{angry} and \emph{neutral} into \emph{sad}.
The latter is particularly problematic as more \emph{neutral} samples are classified as \emph{sad} ($1596$) than \emph{neutral} ($1216$).

These problems are largely mitigated through the use of multimodal architectures.
The improvements introduced by both SFCNN14 and MFCNN14 are mostly concentrated on the \emph{happy} and \emph{neutral} classes, where \ac{TPR} improves by $+33/39\,\%$ and $+47/27\,\%$, respectively.
However, the two architectures exhibit a different behaviour on their off-diagonal entries.
MFCNN14 substantially worsens the false positives on the \emph{happy} class, with a large increase on the amount of \emph{angry} ($+35\,\%$), \emph{neutral} ($+84\,\%$), and \emph{sad} ($+122\,\%$) samples misclassified as \emph{happy}.
In contrast, SFCNN14 exhibits only a minor deterioration ($+13\,\%$) on \emph{neutral} to \emph{happy} misclassifations.
This illustrates that, although the \ac{UAR} of both models, as shown in \cref{table:emotion}, is comparable for MSP-Podcast, the more granular view provided by the confusion matrices clearly positions SFCNN14 as the winning architecture for this experiment.

Overall, our results clearly demonstrate that the proposed deep fusion methods can lead to substantial gains compared to their baseline, unimodal counterparts.
MFCNN14 shows a more robust behaviour with respect to the different datasets, whereas SFCNN14 shows more desirable properties in terms of confusion matrices.

\section{Emotional dimensions}
\label{sec:dimensions}
After evaluating our methods on categorical emotion recognition, we proceed with modelling emotional dimensions.
As discussed in \cref{sec:setup}, we begin with single-task models trained with an \ac{MSE} loss in \cref{subsec:singletask}, which allows us to study the effects of fusion independently for each dimension.
Then, in \cref{subsec:multitask}, we investigate the combination of our methods with multi-task training and a \ac{CCC} loss, which, in line with previous works, enables us to obtain better performance.
\subsection{Single-task models}
\label{subsec:singletask}
\begin{table}
    \caption{
    \Ac{CCC}/\ac{PCC} in-/cross-domain results for emotional dimension prediction using single-task models trained with an \ac{MSE} loss.
    MSP-Podcast in- and cross-domain results computed on the test set, whereas IEMOCAP in-domain results correspond to \acl{LOSO} \acl{CV} and cross-domain results reported on the entire dataset.
    Average (and standard deviation) results computed over 5 runs.
    Fusion results (SFCNN14 and MFCNN14) that are significantly different than the unimodal baseline (CNN14) as determined by two-sided independent sample t-tests ($p < 0.05$) are marked by $^*$.
	}
	\label{table:dimensions:single}
	\centering
	\scriptsize
    \begin{tabular}{c|lll|lll}
    \toprule
    \textbf{Train} & \multicolumn{6}{c}{\textbf{MSP-Podcast}}\\
    \midrule
    \textbf{Test} & \multicolumn{3}{c|}{\textbf{MSP-Podcast}} & \multicolumn{3}{c}{\textbf{IEMOCAP}}\\
    \midrule
    & \textbf{Arousal} & \textbf{Valence} & \textbf{Dominance} & \textbf{Arousal} & \textbf{Valence} & \textbf{Dominance}\\
    \midrule
    & \textbf{\ac{CCC}} & \textbf{\ac{CCC}} & \textbf{\ac{CCC}} & \textbf{\ac{PCC}} & \textbf{\ac{PCC}} & \textbf{\ac{PCC}}\\
    \midrule
    \textbf{CNN14} & \textbf{.664 (.036)} & .217 (.010) & \textbf{.583 (.022)} & \textbf{.593 (.009)} & .323 (.018) & \textbf{.471 (.019)}\\
    \textbf{SFCNN14} & .620 (.025) & .367 (.018)$^*$ & .523 (.019)$^*$ & .543 (.019)$^*$ & .328 (.026) & .417 (.014)$^*$\\
    \textbf{MFCNN14} & .627 (.010) & \textbf{.407 (.009)}$^*$ & .539 (.011)$^*$ & .558 (.015)$^*$ & \textbf{.381 (.017)}$^*$ & .470 (.017)\\
    \midrule
    \textbf{Train} & \multicolumn{6}{c}{\textbf{IEMOCAP}}\\
    \midrule
    \textbf{Test} & \multicolumn{3}{c|}{\textbf{IEMOCAP}} & \multicolumn{3}{c}{\textbf{MSP-Podcast}}\\
    \midrule
    & \textbf{Arousal} & \textbf{Valence} & \textbf{Dominance} & \textbf{Arousal} & \textbf{Valence} & \textbf{Dominance}\\
    \midrule
    & \textbf{\ac{CCC}} & \textbf{\ac{CCC}} & \textbf{\ac{CCC}} & \textbf{\ac{PCC}} & \textbf{\ac{PCC}} & \textbf{\ac{PCC}}\\
    \midrule
    \textbf{CNN14}   & .618 (.010) & .385 (.015) & .424 (.020) & .418 (.008) & .087 (.010) & \textbf{.438 (.007)} \\
    \textbf{SFCNN14} & .551 (.019)$^*$ & .622 (.007)$^*$ & .438 (.013) & .271 (.017)$^*$ & .212 (.004)$^*$ & .210 (.024)$^*$\\
    \textbf{MFCNN14} & \textbf{.628 (.005)} & \textbf{.664 (.005)}$^*$ & \textbf{.503 (.006)}$^*$ & \textbf{.432 (.006)}$^*$ & \textbf{.219 (.005)}$^*$ & .365 (.008)$^*$ \\
    \bottomrule
    \end{tabular}
\end{table}

Our first experiments are performed on the emotional dimensions of MSP-Podcast and IEMOCAP.
In \cref{table:dimensions:single}, we report \ac{CCC} and \ac{PCC} results for in-domain and cross-domain performance, respectively.
The performance of both fusion models is compared to that of the baseline CNN14 using two-sided independent sample t-tests.

The best performance on valence, both in- and cross-domain is achieved by MFCNN14, which reaches a mean \ac{CCC} of $.407$ on MSP-Podcast and $.664$ on IEMOCAP.
This is significantly higher than CNN14 and considerably outperforms SFCNN14, showing that multistage fusion can better utilise the textual information in this case.
Cross-domain performance is severely degraded when training on IEMOCAP and testing on MSP-Podcast, while not so much when doing the opposite.
This illustrates how training on large, naturalistic corpora leads to better generalisation for \ac{SER} systems, both unimodal and bimodal ones.

In the case of arousal, CNN14 performs better on MSP-Podcast with an average \ac{CCC} of $.664$ (vs $.620$ and $.627$ for SFCNN14 and MFCNN14), but this difference is not statistically significant.
On IEMOCAP, MFCNN14 shows a marginally better performance than CNN14, while SFCNN14 is significantly worse than its unimodal baseline.
This curious case shows how additional information can also hamper the training process. 
We hypothesise that this is because textual information is not conducive to arousal modelling, and leads it to perform worse on this task.
This is corroborated by BERT models trained to jointly predict arousal/valence/dominance presented in \cref{subsec:multitask}.
As mentioned in \cref{sec:setup}, we did not train BERT models for each dimension in isolation to reduce the computational load of our experiments; thus, we return to this point in \cref{subsec:multitask}.

Finally, we note that cross-domain performance for arousal, while also lower than in-domain performance, is not as low as valence, especially for CNN14.
Interestingly, \ac{PCC} on IEMOCAP for CNN14 models trained on MSP-Podcast is now significantly higher than the \ac{PCC} obtained by either SFCNN14 or MFCNN14 ($.593$ vs $.543$ and $.558$, respectively).
On the contrary, MFCNN14 shows significantly better performance on the opposite setup ($.432$ vs $.418$ \ac{PCC}) than CNN14, while SFCNN14 remains significantly worse.
This illustrates once more that the paralinguistic information stream carries more information on arousal than the linguistic one, and models trained on that can generalise better across different datasets.

Results on dominance follow the trends exhibited by arousal.
CNN14 is significantly better than SFCNN14 and MFCNN14 on in-domain MSP-Podcast results ($.583$ vs $.523$ and $.539$), but, in the case of MFCNN14, this large in-domain difference does not translate to better cross-domain generalisation, as both models are nearly equivalent on IEMOCAP \ac{PCC} performance ($.471$ vs $.470$).
This tendency is reversed on IEMOCAP; there MFCNN14 achieves significantly better in-domain results ($.503$ vs $.424$), but shows evidence of overfitting by performing significantly worse cross-domain ($.365$ vs $.438$).

Overall, our results show that bimodal fusion significantly improves performance on the valence dimension both in- and cross-domain for both datasets, with MFCNN14 achieving consistently superior performance to SFCNN14.
For the case of MSP-Podcast, the other two dimensions fail to improve, while for IEMOCAP they improve only in-domain, and only for MFCNN14, while SFCNN14 performs consistently worse than CNN14.
As we discuss in \cref{subsec:multitask}, this is because BERT is not good at modelling arousal and dominance, and this propagates to the fusion models.
It thus appears that linguistic information, which by itself is not adequate to learn the tasks, hampers the training process and results in worse fusion models as well.
This undesirable property seems to affect SFCNN14 more strongly than MFCNN14, which is able to circumvent it, and, in some cases, benefit from linguistic information.
Thus, in the case of dimensional emotion recognition, MFCNN14 so far shows a better behaviour than SFCNN14.

\subsection{Multi-task models}
\label{subsec:multitask} 
\begin{table}[t]
    \centering
    \caption{
    \Ac{CCC} results for emotional dimension prediction using multi-task models on MSP-Podcast trained with a \ac{CCC} loss.
    Models trained to jointly optimise the \ac{CCC} for all dimensions.
    Average (and standard deviation) results reported over 5 runs.
    Fusion results (SFCNN14 and MFCNN14) that are significantly different than the unimodal baselines for each dimension as determined by two-sided independent sample t-tests ($p < 0.05$) are marked by $^*$ (for CNN14) and $^\dag$ (for BERT).
    }
    \label{table:affect:multitask:all}
    \begin{tabular}{c|lll}
        \toprule
        \textbf{Architecture} & \textbf{Arousal} & \textbf{Valence} & \textbf{Dominance}\\
        \midrule
        & \textbf{\ac{CCC}} & \textbf{\ac{CCC}} & \textbf{\ac{CCC}}\\
        \midrule
        \textbf{BERT} & .232 (.006) & .503 (.003) & .214 (.008) \\
        \textbf{CNN14} & .660 (.012) & .291 (.029) & .578 (.011)\\
        \textbf{SFCNN14} & .665 (.008)$^\dag$ & .497 (.013)$^{*\dag}$ & .598 (.025)$^\dag$\\ 
        \textbf{MFCNN14} & \textbf{.678 (.005)$^{*\dag}$} & \textbf{.521 (.004)}$^{*\dag}$ & \textbf{.604 (.001)$^{*\dag}$}\\
        \bottomrule
    \end{tabular}
\end{table}

We end our section on emotional dimensions by considering a multi-task problem with a \ac{CCC} loss.
This is motivated by several recent works who have gotten better performance by switching to this formulation~\citep{parthasarathy2017jointly,li2021contrastive}.
To reduce the footprint of our experiments, we only evaluate this approach on MSP-Podcast.
\ac{CCC} results for $5$ runs are shown in \cref{table:affect:multitask:all}.
As previously discussed, \cref{table:affect:multitask:all} additionally includes results with a fine-tuned BERT model.
As expected, we observe that BERT performs much better than CNN14 on valence prediction ($.503$ vs $.291$), but lacks far behind on arousal ($.232$ vs $.660$) and dominance ($.214$ vs $.578$).
This clearly illustrates how the two streams, acoustics and linguistics, carry complementary information for emotion recognition.

Both fusion methods improve on all dimensions compared to CNN14.
In particular, MFCNN14 is significantly better on all three dimensions ($.678$ vs $.660$, $.521$ vs $.291$, and $.604$ vs $.578$), whereas SFCNN14 is significantly better only for valence ($.497$ vs $.291$) but not for arousal ($.665$ vs $.660$) and dominance ($.598$ vs $.578$).
Moreover, of the two fusion methods, only MFCNN14 significantly improves on valence performance compared to BERT ($.521$ vs $.503$), while SFCNN14 performs marginally (but significantly) worse ($.497$).
This demonstrates once more that multistage fusion can better utilise the information coming from the two streams.

\begin{figure*}[!t]
    \centering
    \includegraphics[width=\textwidth]{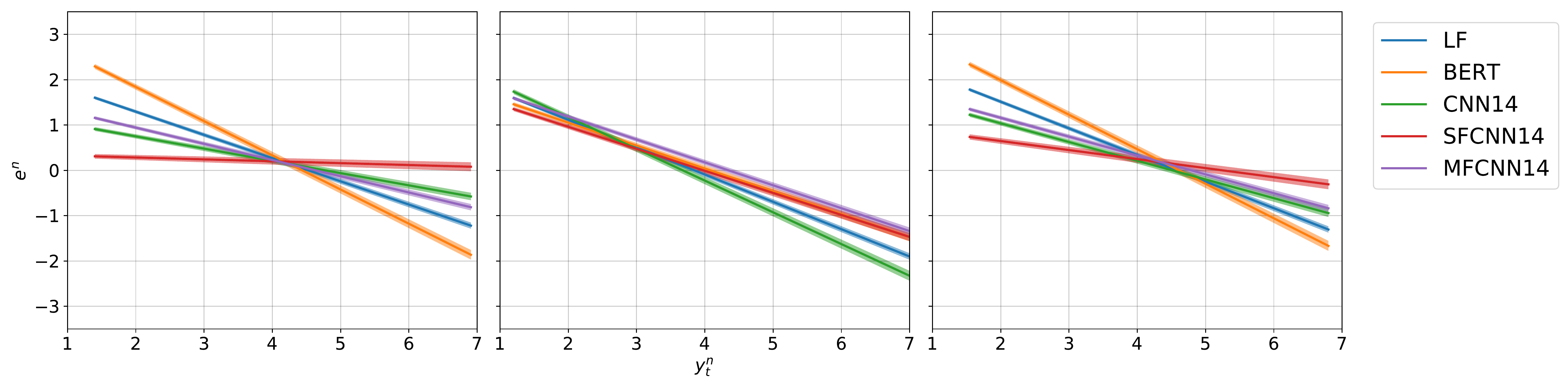}
    \caption{
        Error residual plots for arousal (left), valence (middle), and dominance (right).
        Linear curves fitted on error residuals ($e^n = y^n_t - y^n_p$) for each architecture and dimension and plotted against the gold standard ($y^n_t$).
        With the exception of SFCNN14 on arousal, all models show higher errors towards the edges of their scale.
        Plots best viewed in colour.
    }
   \label{fig:residuals}
\end{figure*}
Finally, we are interested in whether the models show a heteroscedastic behaviour by examining their error residuals.
To this end, we pick the best models on the validation set and examine their residuals (model results for the selected seeds included in \ref{app:residuals}).
\cref{fig:residuals} shows fitted linear curves on the error residuals for each model and task.
We use fitted curves for illustration purposes as superimposing the scatterplots for each model would make our plots uninterpretable.
The curves are least-squares estimates over all error residuals.
Our analysis reveals that most models show non-uniform errors, with their deviation from the ground truth increasing as we move away from the middle of the scale.
This `regression towards the mean' phenomenon is highly undesirable, especially for real-world applications where users would observe a higher deviation from their own perception of a target emotion for more intense manifestations of it.

SFCNN14 is the only model which escapes this undesirable fate, primarily for arousal and dominance.
BERT, in contrast, shows the worst behaviour for those two dimensions, but is comparable to SFCNN14 for valence. 
CNN14 and MFCNN14 show improved a much behaviour compared to BERT, but some bias still appears, with the late fusion baseline naturally falling in the middle between BERT and CNN14.
For valence, SFCNN14 and MFCNN14 both closely follow BERT in showing a low, but nevertheless existent bias, while CNN14 performs worse.

Interestingly, the residuals are also showing an asymmetric behaviour.
For valence, in particular, CNN14 is showing higher errors than the other models for the upper end of the scales, but is comparable to them for the lower end.
Conversely, BERT shows higher errors for the lower end of arousal and dominance.
This indicates that models struggle more with the scales.
Naturally, this is partly explained by the sparseness of data for the more extreme values, as naturalistic datasets tend to be highly imbalanced towards neutral.
Nevertheless, this continues to pose a serious operationalisation problem for \ac{AER} systems.

\section{Conclusion}
\label{sec:conclusion}

In the present contribution, we investigated the performance of deep fusion methods for emotion recognition.
We introduced a novel method for combining linguistic and acoustic information for \ac{AER}, relying on deep, multistage fusion of summary linguistic features with the intermediate layers of a \ac{CNN} operating on log-Mel spectrograms, and contrasted it with a simpler, single-stage fusion one where information is only combined at a single point.
We demonstrated both methods' superiority over unimodal and shallow, decision-level fusion baselines.
In terms of quantitative evaluations, multistage fusion fares better than the single-stage baseline, thus illustrating how a tighter coupling of acoustics and linguistics inside \acp{CNN} can lead to a better integration of the two streams.



\appendix

\section{Error residuals}
\label{app:residuals}
Results for these the best-performing multitask emotional dimension prediction models are shown in \cref{table:affect:multitask:best}.
As we are using error residuals for our evaluation, we additionally show \ac{MSE} results.

\begin{table}[!ht]
    \centering
    \caption{
    \Ac{CCC} and \ac{MSE} results for emotional dimension prediction using multi-task models on MSP-Podcast.
    Models trained to jointly optimise the \ac{CCC} for all dimensions.
    Results reported for the model showing the best validation set performance for each architecture.
    LF corresponds to a late, decision-level fusion (averaging) of unimodal predictions.
    }
    \label{table:affect:multitask:best}
    \small
    \begin{tabular}{c|llllll}
        \toprule
         & \multicolumn{2}{c}{\textbf{Arousal}} & \multicolumn{2}{c}{\textbf{Valence}} & \multicolumn{2}{c}{\textbf{Dominance}}\\
        \midrule
        \textbf{Architecture} & \textbf{\ac{CCC}} & \textbf{\ac{MSE}} & \textbf{\ac{CCC}} & \textbf{\ac{MSE}} & \textbf{\ac{CCC}} & \textbf{\ac{MSE}}\\
        \midrule
        \textbf{BERT}    & .228 & 1.352 & .462 & 1.027 & .209 & 0.893\\
        \textbf{CNN14}   & .658 & 0.617 & .248 & 1.625 & .564 & 0.449\\
        \textbf{LF}      & .574 & 0.587 & .460 & \textbf{0.832} & .496 & 0.414\\
        \textbf{SFCNN14} & .657 & 0.813 & .481 & 0.990 & .561 & 0.620\\
        \textbf{MFCNN14} & \textbf{.670} & \textbf{0.514} & \textbf{.515} & \textbf{0.832} & \textbf{.603} & \textbf{0.381}\\
        \bottomrule
    \end{tabular}
\end{table}

\section{\refname}
\printbibliography[heading=none]

\end{document}